%% file: acl2019.tex
\documentclass[11pt,a4paper]{article}
\usepackage[hyperref]{acl2019}
\usepackage{times}
\usepackage{xcolor}
\usepackage{latexsym}
\usepackage{amsmath}
\usepackage{graphicx}
\usepackage{booktabs}
\usepackage{colortbl}

\usepackage{url}

\aclfinalcopy 

\title{Toward Dialogue Modeling: A Semantic Annotation Scheme for Questions and Answers}

\author{
 Mar\'ia Andrea Cruz Bland\'on, Gosse Minnema, Aria Nourbakhsh\\ \textbf{Maria Bortichev, Maxime Amblard} \\
 LORIA, UMR 7503, Universit́\'e de Lorraine, CNRS, Inria\\
 Nancy, France \\
 {\tt \{mariaandrea.cruzblandon, gosseminnema\}@gmail.com}\\
 {\tt aria.nourbakhsh@outlook.com} \\
{\tt \{maria.boritchev, maxime.amblard\}@univ-lorraine.fr}}

\date{}

\begin{document}
\maketitle

\begin{abstract}

The present study proposes an annotation scheme for classifying the content and discourse contribution of question-answer pairs. We propose detailed guidelines for using the scheme and apply them to dialogues in English, Spanish, and Dutch. Finally, we report on initial machine learning experiments for automatic annotation. 

\end{abstract}

\newcommand{\misscite}{\textcolor{purple}{(missing citation)}}

\input{introduction.tex}

\input{related_work.tex}
\input{annotation_scheme.tex}
\input{annotation_experiments.tex}

\input{conclusions.tex}

\section*{Acknowledgements}

This paper was written while the first authors (Cruz Bland\'on, Minnema, Nourbakhsh) were enrolled in the European Master Program in Language and Communication Technologies (LCT) and were supported by the European Union Erasmus Mundus program. 

\bibliography{acl2019}
\bibliographystyle{acl_natbib}

\end{document}

%% file: introduction.tex
\section{Introduction}

Question-answer pair (QAP) labeling is the problem of characterizing the content and discourse contribution of questions and answers using a small but maximally informative tagset that can be consistently applied by both human annotators and NLP systems. QAP labeling has many potential use cases, for example as a preprocessing step for dialogue modeling systems or for chatbots. The problem is not new: in the NLP literature, different aspects of QAP tagging have been addressed in the context of question answering systems \cite{li-roth2002}, question generation systems \cite[e.g.][]{graesser-etal2008}, and dialogue act classification \cite[e.g.][]{allen-core1997, stolcke-etal2000}. 

However, we see several gaps in the literature: existing approaches to QAP classification often do not cover the full range of questions and answers found in human dialogues 
and are limited in the types of semantic information that they cover. To address these issues, we propose a new annotation scheme that was developed based on corpora of natural conversations in several languages (English, Spanish, and Dutch) and provides several layers of annotations for QAPs. Notably, where applicable, we annotate the semantic role of the questioned constituent in questions and their corresponding answer (e.g. `Does she live in \textit{Paris or London?}' $\Rightarrow$ \textsc{location}), which we believe is an informative, yet easy definable way of globally characterizing the content of a QAP. 

Our paper has two main contributions: the annotation scheme itself (section \ref{sec:scheme}) and two ways of applying it to real data. We developed detailed and explicit guidelines for human annotators, and tested these on corpus data (section \ref{sub:manual}). Additionally, we started experimenting with machine learning approaches for automating part of the annotation process (section \ref{sub:ml}).

%% file: related_work.tex
\section{Related Work}
\label{sec:related}

Our annotation scheme is related to two existing schemes in particular. The first of these is \citet{FREED1994621}, which categorizes questions along an \textit{information continuum} that ranges from questions purely asking for factual information to questions that convey, rather than request, (social) information. Within this continuum, questions are divided into classes that are defined based on a combination of formal (syntactic) and functional criteria. Both of these ideas are also used in our scheme: our question types are also distinguished by whether they ask or convey information (`phatic questions' and `completion suggestions' fall into the latter category) and are defined as combinations of specific forms and functions. 


Another related scheme is \citet{stolcke-etal2000}, an adapted version of DAMSL (`Dialog Act Markup in Several Layers', \citealt{allen-core1997}), an annotation scheme for dialogue acts (including QAPs). The scheme includes a set of eight different question types (e.g. \textit{yes/no} questions, \textit{wh-}questions, \textit{rhetorical} questions) that has considerable overlap with our set of question types. 

%% file: annotation_scheme.tex
\section{Annotation scheme}
\label{sec:scheme}

Annotated information is split between two main `layers': \textit{question/answer type} and \textit{feature} (semantic role). Every question or answer is assigned at least a type tag, and depending on the type, a feature tag. 



\subsection{Questions}

The question tagset was designed in a corpus-driven way, starting with two basic types and expanding the tagset based on corpus data. Our starting assumption is that the corpora would contain at least two well-known and well-defined categories of questions:
\textit{yes/no} questions and \textit{wh}-questions \citep{FREED1994621}. In our opinion, both of these types are useful \textit{a priori}, because they are each associated with a clear set of syntactic, semantic, and pragmatic 
characteristics (at least for the languages that are included in this study). Prototypical English \textit{yes/no} questions are characterized by subject-auxiliary inversion and do-support (syntax), express a proposition that could be true or false (semantics), and their answers are expected to either confirm or deny this proposition (pragmatics). On the other hand, a prototypical English \textit{wh}-question contains a fronted constituent that starts with a \textit{wh}-word (syntax), expresses a proposition with missing information (semantics), and expects the answerer to supply this missing information (pragmatics)  \citep{FREED1994621}.

Next, we looked for questions in our corpora that did not correspond to either of the two prototypes and extended the scheme to fit them (see table \ref{tab-question_types} for the final scheme and examples).
First, there are questions that are similar to \textit{wh-}questions or \textit{yes/no} questions but have a deviant form (e.g. \textit{wh-in-situ} questions like `You saw what?', or yes/no questions without inversion such as `You saw him?'). We decided not to introduce new categories for these on the basis of their semantics and pragmatics.

A second group of questions has the syntactic characteristics of a \textit{yes/no} question or a \textit{wh}-question, but a different pragmatics and/or semantics. For example, the asker of the question suggests a way to complete the utterance of the previous speaker, and the expected answer would confirm or deny this suggestion. This is subtly different from a prototypical \textit{yes/no} question because the asker of the question
does not necessarily 
ask their interlocutor to confirm the truth value of the suggestion (e.g. \textit{A: it includes heat and uhm, I think B: Water?}, SCoSE/Amy, line 746-747\footnote{See section \ref{sub:corpora} for information about our corpora.}). We call these types of questions \textit{completion suggestions}.


\begin{table}[h]
\centering
\resizebox{0.48\textwidth}{!}{%
\begin{tabular}{@{}ll|ll@{}}
\toprule
\textbf{Tag} & \textbf{Name} & \textbf{Tag} & \textbf{Name}    \\ \midrule
YN  & Yes/No question  & WH  & Wh-question \\
CS  & Completion suggestion & PQ & Phatic question \\
DQ  & Disjunctive question  \\
\bottomrule
\end{tabular}
}
\caption{Question types}
\label{tab-question_types}
\end{table}

The third group of questions appear to be a \textit{yes/no} question or a \textit{wh}-question, respectively, but their context and intonation make clear that the asker is not actually interested in the confirmation or denial of the proposition. Instead, such questions can have various so-called \textit{phatic functions}, i.e. their semantic content is less important than their social and rhetorical functions \citep{FREED1994621, senft2009}. We call this type of questions \textit{phatic questions} (e.g. \textit{right?} / \textit{oh yeah?} / \textit{you know?}).\footnote{Note that our use of the term \textit{phatic question} is somewhat broader than the \textit{phatic information} question described in \citet{FREED1994621}; for example, our definition also includes rethorical questions, while in Freed's scheme, these are not included.}

Finally, some questions containing a disjunction (e.g. `Do you go on Monday or on Tuesday?') are semantically and pragmatically similar to \textit{wh}-questions, but are syntactically closer to  \textit{yes/no} questions. This kind of questions, like \textit{yes/no} questions, exhibits subject-auxiliary inversion (at least in English), but does not ask for the confirmation or denial of the proposition that it expresses. Instead, it expects the answerer to provide some missing information with the set of options to choose from. We call this type of questions \textit{disjunctive questions} (sometimes also called \textit{alternative questions} in the literature).


\subsection{Features}

\textit{Wh-} and \textit{disjunctive} questions are always `about' a particular constituent (e.g. `\textit{Which man} is running?', `Do you want \textit{coffee or tea}?'). The \textit{feature}, or semantic role of this constituent provides information about the content of the question and the expected answer (e.g. if the questioned constituent is an \textsc{agent} then it is likely that the answer will refer to a person). Detecting semantic roles requires semantically analyzing the sentence, but for \textit{wh-}questions, \textit{wh}-words often provide cues (e.g. `where' for \textsc{location}). Our feature annotations follow the feature set (see table \ref{tab-Features}) and the mapping from (English) \textit{wh}-words to features proposed in \citet{boritchev2017} (adapted from \citealt{Jurafsky:2000:SLP:555733}).

\begin{table}[h]
\centering
\begin{tabular}{@{}ll|ll@{}}
\toprule
\textbf{Tag} & \textbf{Name} & \textbf{Tag} & \textbf{Name} \\ \midrule
TMP          & Temporality    & OW           & Owner          \\
LOC          & Location       & RE           & Reason         \\
AG           & Agent          & TH           & Theme          \\
CH           & Characteristic \\ 
\bottomrule
\end{tabular}
\caption{Features}
\label{tab-Features}
\end{table}

\subsection{Answers}

The main intuition underlying our answer annotation scheme is that question types restrict their answers: for example, \textit{yes/no} questions are prototypically answered by `yes' or `no', and \textit{wh}-questions ask for a constituent with a particular feature. Table \ref{tab-answers} summarizes our answer types and their corresponding question types. Among these types of answers, there may be overlaps. For example, a `deny the assumption' answer can be thought of as a negative answer because it is possible that they share the same grammatical and semantic structure. Different factors including the context and prosody are relevant to decide between overlapping tags.



Some questions are not followed by answer. We distinguish between two situations. First, there are questions that receive a reply that, while not providing the information asked for in the question, clearly do respond to it. For example, in the QAP \textit{A: `When will you guys get off?' / B: `My last exam is like \ldots I don't know'} (SCoSe/Amy, line 243-244), B's response does not answer A's question directly but does engage with it as there is a logical connection between finishing the exams and going on vacation. In such cases, the response is tagged as \textit{unrelated topic (UT)} because it is about a different topic but still responds to the question. By contrast, when there is no response at all, no answer should be annotated.


\begin{table}[h!]
\resizebox{0.48\textwidth}{!}{%
\begin{tabular}{@{}lll@{}}
\toprule
\textbf{Tags} & \textbf{Name}       & \textbf{Question Type} \\ \midrule
PA            & Positive Answer     & YN, CS                 \\
NA            & Negative Answer     & YN, CS                 \\
FA            & Feature Answer      & DQ, WH                 \\
PHA           & Phatic Answer       & YN, CS, DQ, WH, PQ     \\
UA            & Uncertainty Answers & YN, CS, DQ, WH, PQ     \\
UT            & Unrelated Topic     & YN, CS, DQ, WH, PQ     \\
DA            & Deny the Assumption & YN, CS, DQ, WH, PQ     \\ \bottomrule
\end{tabular}
}
\caption{Answers}
\label{tab-answers}
\end{table}

%% file: annotation_experiments.tex
\section{Annotation Experiments}
\label{sec:experiments}

In this section, we discuss our experiments with applying the scheme manually (section \ref{sub:manual}) and using machine learning techniques (section \ref{sub:ml}).

\input{manual_annotations.tex}
\input{machine_learning.tex}

%% file: manual_annotations.tex
\subsection{Manual annotation}
\label{sub:manual}

We have experimented with applying the scheme on real-world data. Our experiment consists of two parts: writing annotation guidelines to explicitly define the annotation process and annotating 701 questions across three languages, namely, English, Spanish, and Dutch.\footnote{Our guidelines and annotations are available in our repository at \url{https://github.com/andrea08/question_answer_annotation}.} 

\subsubsection{Annotation guide}
In order to help annotators apply the scheme consistently, we wrote annotation guidelines for English, which include examples and instructions for how to use the annotation software~(ELAN \citeyear{ELAN}, \citealt{Sloetjes2008AnnotationBC}). The annotation procedure guides the annotator in identifying questions, dealing with transcription errors, determining question types, and adding tags for additional information such as features, complexity, and indirectness. 

Some question types have a very specific prototypical syntactic form (e.g. \textit{wh}-questions), whereas other questions can have several different forms (e.g. \textit{phatic} questions). We exploit this by defining a precedence order for question types, which serves as a filter for identifying questions. The precedence order lists question types from the most specific to the most general ones, i.e. from questions with easily identifiable characteristics to those that can have different forms as it is the case for the phatic questions. The  precedence order is as follows: (1) \emph{Wh-questions}, (2) \emph{Disjunctive questions}, (3) \emph{Yes/No questions}, (4) \emph{Completion suggestions} (5) \emph{Phatic questions.}

\subsubsection{Corpora}
\label{sub:corpora}

We annotated several dialogues from three different corpora in three languages: the \textit{Saarbr\"ucken Corpus of Spoken English (SCoSE)} \citep{SCoSE}, a corpus of face-to-face conversations; the \textit{CallFriend} corpus (Spanish)  \citep{CallFriend}, a corpus of phone conversations; and the \textit{Spoken Dutch Corpus (CGN)} \citealt{CGN}, a corpus of phone conversations. The purpose of annotating these dialogues was to test the annotation scheme on different languages and produce annotated data. 

We annotated all questions in a subset of 4,939 utterances from the SCoSE corpus. Of these, 3,578 utterances were used to build the `gold standard' corpus (used for calculating agreement scores and training machine learning algorithms). The remainder of the corpus was used as a test set in the machine learning algorithms. Furthermore, we annotated questions and answers from 2,618 and 935 utterances of CallFriend and CGN corpora, respectively.
We relied primarily on the transcriptions of the corpora; in case of doubt, we made use of the audio recordings as well.

\subsubsection{Results}
We annotated 701 questions (Q) and 483 answers (A), distributed as follows: 422 (Q) / 289 (A) in the ScoSE corpus; 87 (Q) / 72 (A) in the CGN corpus; and 192 (Q) / 122 (A) in the CallFriend corpus. A descriptive analysis of our annotations shows that \textit{yes/no} questions are the most common type in the three corpora, $40\%$ (Spanish), $42\%$ (English) and $64\%$ (Dutch). 

\begin{table}[t!]
\begin{center}

\begin{tabular}{@{}lll@{}}
\toprule
\textbf{Annotators} & \textbf{$A_o$} & \textbf{$\kappa$} \\ \midrule
Questions           & 0.73        & 0.63       \\
Features            & 0.90        & 0.67       \\
Answers             & 0.59        & 0.49       \\ \bottomrule
\end{tabular}
\end{center}
\caption{\label{tab:kappa-table} Cohen's Kappa score ($\kappa$) and observed agreement ($A_{o}$) for gold standard dialogue}
\end{table}

To evaluate the annotations, inter-annotator agreement was calculated based on a subset of the gold standard corpus.\footnote{This subset consists of the 690 utterances jointly annotated by all three annotators.} Table \ref{tab:kappa-table} illustrates the values of observed agreement ($A_{o}$) and Cohen's $\kappa$ \citep{doi:10.1177/001316446002000104} obtained for question, feature and answer annotation. 
The agreement values obtained for question types were over $0.6$ (for all annotators combined). This would generally be considered to be a `moderate' level of agreement \citep{10.2307/2529310}. A large share of our disagreements came from phatic questions; distinguishing these from other question types sometimes relies on subtle pragmatic and semantic contextual judgements. 
Agreement for answer types is lower than for question types because question types restrict answer types and hence question type disagreements can cause answer type disagreements.

In order to improve the annotation guidelines, we systematically examined all of the disagreements, 
most of which fell into one of four categories: 
(1) Simple mistakes, such as missing a question or choosing an (obviously) wrong tag. 
(2)~Disagreements as a consequence of a previous disagreement; e.g., \textit{wh}-questions need feature annotations, but \textit{phatic} questions do not. In this case, a disagreement about the question type can cause further disagreement about feature type. (3)~Missing instructions in the annotation guidelines for handling particular situations, e.g.
annotating utterances containing interruptions. 
(4)~ Utterances whose interpretation was ambiguous and depends on subtle intonational or contextual cues for which it is hard to formulate a general rule.

%% file: machine_learning.tex
\subsection{Machine learning}
\label{sub:ml}

We also conducted preliminary machine learning experiments for automating the annotation process. For the moment, we focus only on question type classification for English dialogues. So far, the approach that shows the most promising results is a decision tree algorithm \cite{quinlan1986induction} that takes as input a set of hand-designed features representing formal characteristics of a question, such as its length, the presence of a \textit{wh}-word, and the presence of words such as \textit{really?} or \textit{you know?} Our full feature set is given in Table \ref{tab-feature_extraction}. Note that these features are quite superficial and do not take into account the discourse context of a question. Still, the algorithm achieves an accuracy score of $0.73$ and an F1-score of $0.58$, outperforming our majority-class baseline algorithm by a wide margin ($acc.=0.47$, $F1=0.31$).\footnote{A global $F1$ score was calculated by macro-averaging the scores for individual classes.} 

Analysing the effect of the features in the predictions of the decision tree, we found that the majority of the mistakes were associated with the length of the questions. From the questions that were misclassified and had a length less than $6$ ($26$ questions), $50\%$ were wrongly predicted as \textit{phatic} questions. Particularly, as with manual annotations, \textit{phatic} questions that contain \textit{wh-words} were source of disagreement and misclassified. 
Table \ref{tab:confusion_matrix} shows the confusion matrix for all the question types. 

\begin{table}[h]
\renewcommand{\arraystretch}{1.3}
\centering
\resizebox{0.47\textwidth}{!}{
\begin{tabular}{@{}p{1.6cm}p{5cm}l@{}}
\toprule
\textbf{Feature}      & \textbf{Description}                               & \textbf{Value}  \\ \midrule
has\_wh               & Contains a wh-constituent                   & True, False     \\
has\_or               & Contains the word ``or''                              & True, False     \\
has\_\allowbreak inversion        & Verb before NP (based on shallow parse)                                 & True, False     \\
has\_tag              & Contains a tag (`isn't it', `right')                     & True, False     \\
last\_utt\_\allowbreak similar     & Question shares $\geq$ 50\% of its words with the previous utterance          & True, False     \\
last\_utt\_\allowbreak incomplete & Previous utterance is interrupted (marked with special transcription symbol)                   & True, False     \\
has\_cliche           & Contains a phatic marker (`you know?', `really?') & True, False     \\
length                & Number of words       & Numerical \\ \bottomrule
\end{tabular}%
}
\caption{Extracted features for the classification task}
\label{tab-feature_extraction}
\end{table}

\begin{table}[h]
\centering
\resizebox{0.4\textwidth}{!}{
\begin{tabular}{@{}lrrrrrr@{}}
\toprule
 & \multicolumn{1}{l}{\textbf{YN}} & \multicolumn{1}{l}{\textbf{DQ}} & \multicolumn{1}{l}{\textbf{PQ}} & \multicolumn{1}{l}{\textbf{CS}} & \multicolumn{1}{l}{\textbf{WH}} & \multicolumn{1}{l}{\textbf{Support}} \\ \midrule
\textbf{YN} & 74 & 1 & 8 & 3 & 2 & 88 \\
\textbf{DQ} & 0 & 3 & 0 & 0 & 0 & 3 \\
\textbf{PQ} & 7 & 0 & 15 & 0 & 8 & 30 \\
\textbf{CS} & 1 & 0 & 0 & 0 & 0 & 1 \\
\textbf{WH} & 10 & 0 & 9 & 0 & 43 & 62 \\ \bottomrule
\end{tabular}%
}
\caption{Confusion matrix of decision tree prediction (test set, 184 questions)}
\label{tab:confusion_matrix}
\end{table}

Furthermore, we experimented with two neural architectures, a bag-of-words (BOW) classifier and a recurrent neural network (RNN), to test what input representations are most informative. However, so far these models suffer from overfitting and perform worse than the decision tree model (BOW: $acc.=0.76$, $F1=0.44$; RNN: $acc.=0.54$, $F1=0.24$). We expect these models to perform better when more training data is available.



%% file: conclusions.tex

\section{Conclusion}
\label{sec:conclusion}

This paper introduced a new annotation scheme for question-answer pairs in natural conversation. The scheme defines five question types and seven answer types based on a mix of formal and functional criteria. An annotation guide was developed and multi-lingual corpora were annotated. Inter-annotator agreement scores were moderately high; a qualitative analysis of disagreements led to improvements to the annotation guidelines.  Initial machine learning experiments show that a simple decision tree algorithm achieves above-baseline performance, but much work remains to be done for making automatic annotation practically feasible. For future work, we would also like to expand the multilingual component of our work by adding language-specific guidelines, annotating more corpora, and adapting our machine learning algorithms to different languages. 

%% file: acl2019.bbl
\begin{thebibliography}{16}
\expandafter\ifx\csname natexlab\endcsname\relax\def\natexlab#1{#1}\fi

\bibitem[{Allen and Core(1997)}]{allen-core1997}
James Allen and Mark Core. 1997.
\newblock Draft of {DAMSL}: Dialog act markup in several layers.
\newblock
  \url{https://www.cs.rochester.edu/research/speech/damsl/RevisedManual/},
  accessed January 22, 2019.

\bibitem[{Boritchev(2017)}]{boritchev2017}
Maria Boritchev. 2017.
\newblock {Approaching dialogue modeling in a dynamic framework}.
\newblock Master's thesis, Universit\'e de Lorraine.

\bibitem[{Canavan and Zipperlen(1996)}]{CallFriend}
Alexandra Canavan and George Zipperlen. 1996.
\newblock \href {https://catalog.ldc.upenn.edu/LDC96S58} {{CALLFRIEND},
  {Spanish-Non-Caribbean Dialect} ({LDC Catalog Number: LDC96S58}).}

\bibitem[{Cohen(1960)}]{doi:10.1177/001316446002000104}
Jacob Cohen. 1960.
\newblock \href {https://doi.org/10.1177/001316446002000104} {A coefficient of
  agreement for nominal scales}.
\newblock \emph{Educational and Psychological Measurement}, 20(1):37--46.

\bibitem[{{ELAN (version 5.2)}(2017)}]{ELAN}
{ELAN (version 5.2)}. 2017.
\newblock The Language Archive, Max Planck Institute for Psycholinguistics,
  \nobreak{Nijmegen}, The Netherlands.
  \url{https://tla.mpi.nl/tools/tla-tools/elan/}.

\bibitem[{Freed(1994)}]{FREED1994621}
Alice~F. Freed. 1994.
\newblock The form and function of questions in informal dyadic conversation.
\newblock \emph{Journal of Pragmatics}, 21(6):621 -- 644.

\bibitem[{Graesser et~al.(2008)Graesser, Rus, and Cai}]{graesser-etal2008}
Art Graesser, Vasile Rus, and Zhiqiang Cai. 2008.
\newblock Question classification schemes.
\newblock In \emph{Proceedings of the Workshop on Question Generation}.

\bibitem[{Jurafsky and Martin(2000)}]{Jurafsky:2000:SLP:555733}
Daniel Jurafsky and James~H. Martin. 2000.
\newblock \emph{Speech and Language Processing: An Introduction to Natural
  Language Processing, Computational Linguistics, and Speech Recognition}, 1st
  edition.
\newblock Prentice Hall PTR, Upper Saddle River, NJ, USA.

\bibitem[{Landis and Koch(1977)}]{10.2307/2529310}
J.~Richard Landis and Gary~G. Koch. 1977.
\newblock The measurement of observer agreement for categorical data.
\newblock \emph{Biometrics}, 33(1):159--174.

\bibitem[{Li and Roth(2002)}]{li-roth2002}
Xin Li and Dan Roth. 2002.
\newblock Learning question classifiers.
\newblock In \emph{COLING '02 Proceedings of the 19th international conference
  on computational linguistics}, pages 1--7.

\bibitem[{Norrick(2017)}]{SCoSE}
Neal Norrick. 2017.
\newblock \href {https://ca.talkbank.org/access/SCoSE.html} {{SCoSE} part 1:
  Complete conversations}.
\newblock English Linguistics, Department of English at Saarland University.

\bibitem[{Oostdijk(2001)}]{CGN}
Nelleke Oostdijk. 2001.
\newblock The design of the {Spoken Dutch Corpus}.
\newblock \emph{Language and Computers}, 36:105--112.

\bibitem[{Quinlan(1986)}]{quinlan1986induction}
J.~Ross Quinlan. 1986.
\newblock Induction of decision trees.
\newblock \emph{Machine learning}, 1(1):81--106.

\bibitem[{Senft(2009)}]{senft2009}
Gunter Senft. 2009.
\newblock Phatic communion.
\newblock In Gunter Senft, Jan-Ola {\"O}stman, and Jef Verschueren, editors,
  \emph{Culture and language use}, pages 226--233. John Benjamins Publishing,
  Amsterdam/Philadelphia.

\bibitem[{Sloetjes and Wittenburg(2008)}]{Sloetjes2008AnnotationBC}
Han Sloetjes and Peter Wittenburg. 2008.
\newblock Annotation by category: {{ELAN and ISO DCR}}.
\newblock In \emph{LREC}.

\bibitem[{Stolcke et~al.(2000)Stolcke, Ries, Coccaro, Shriberg, Bates,
  Jurafsky, Taylor, Martin, Van Ess-Dykema, and Meteer}]{stolcke-etal2000}
Andreas Stolcke, Klaus Ries, Noah Coccaro, Elizabeth. Shriberg, Rebecca. Bates,
  Daniel. Jurafsky, Paul Taylor, Rachel Martin, Carol Van Ess-Dykema, and Marie
  Meteer. 2000.
\newblock Dialogue act modeling for automatic tagging and recognition of
  conversational speech.
\newblock \emph{Computational linguistics}, 26(3):339--373.

\end{thebibliography}
